%% file: main.tex
\documentclass[10pt,twocolumn,letterpaper]{article}

\usepackage[pagenumbers]{cvpr} 

\input{preamble}

\definecolor{cvprblue}{rgb}{0.21,0.49,0.74}
\usepackage[pagebackref,breaklinks,colorlinks,citecolor=cvprblue]{hyperref}

\def\paperID{****} 
\def\confName{CVPR}
\def\confYear{2024}

\usepackage{amsfonts,amssymb}
\usepackage{multirow}
\usepackage{color, colortbl}
\usepackage{xcolor}
\usepackage{graphicx}
\usepackage{pifont}
\usepackage{makecell}

\definecolor{Gray}{gray}{0.93}
\definecolor{Red}{RGB}{255, 46, 23}
\definecolor{Green}{RGB}{0, 171, 79}

\newcommand\blfootnote[1]{%
  \begingroup
  \renewcommand\thefootnote{}\footnote{#1}%
  \addtocounter{footnote}{-1}%
  \endgroup
}

\title{OneLLM: One Framework to Align All Modalities with Language}

\author{Jiaming Han$^{1,2}$, Kaixiong Gong$^{1,2}$, Yiyuan Zhang$^{1,2}$, Jiaqi Wang$^{2}$, Kaipeng Zhang$^{2}$\\Dahua Lin$^{1,2}$, Yu Qiao$^{2}$, Peng Gao$^{2}$, Xiangyu Yue$^{1\dag}$ \vspace{0.2cm}\\
$^1$MMLab, The Chinese University of Hong Kong\\
$^2$Shanghai Artificial Intelligence Laboratory
}

\begin{document}
\maketitle

\blfootnote{$^\dag$ Corresponding author}

\begin{abstract}
Multimodal large language models (MLLMs) have gained significant attention due to their strong multimodal understanding capability. However, existing works rely heavily on modality-specific encoders, which usually differ in architecture and are limited to common modalities. In this paper, we present \textbf{OneLLM}, an MLLM that aligns \textbf{eight} modalities to language using a unified framework. We achieve this through a unified multimodal encoder and a progressive multimodal alignment pipeline. In detail, we first train an image projection module to connect a vision encoder with LLM. Then, we build a universal projection module (UPM) by mixing multiple image projection modules and dynamic routing. Finally, we progressively align more modalities to LLM with the UPM. To fully leverage the potential of OneLLM in following instructions, we also curated a comprehensive multimodal instruction dataset, including \textbf{2M} items from image, audio, video, point cloud, depth/normal map, IMU and fMRI brain activity. OneLLM is evaluated on \textbf{25} diverse benchmarks, encompassing tasks such as multimodal captioning, question answering and reasoning, where it delivers excellent performance. Code, data, model and online demo are available at \url{https://github.com/csuhan/OneLLM}.
\end{abstract}

\begin{figure}
    \centering
    \includegraphics[width=\linewidth]{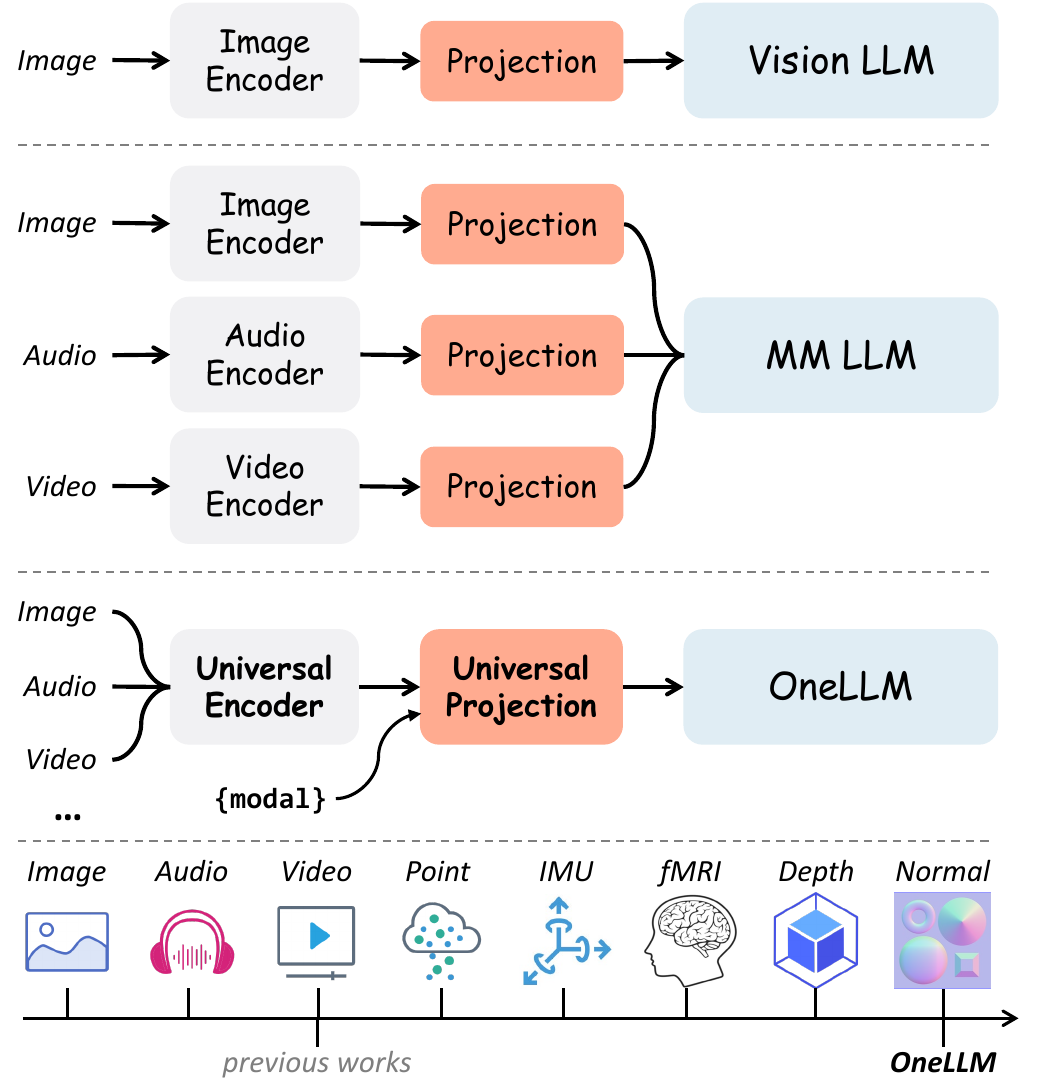}
    \vspace{-7mm}
    \caption{\textbf{Comparisons of Different Multimodal LLMs.} Vision LLM: one image encoder and projection module. Multimodal (MM) LLM: modality-specific encoder and projection module. \textbf{OneLLM}: a universal encoder, a universal projection module and modality tokens $\{\mathrm{modal}\}$ to switch between modalities. \textbf{Bottom:} OneLLM expands supported modalities from three to eight.}
    \vspace{-3mm}
    \label{fig:framework_small}
\end{figure}

\section{Introduction}
\label{sec:intro}

Large Language Models (LLMs) are getting increasingly popular in the research community and industry due to their powerful language understanding and reasoning capabilities. Notably, LLMs such as GPT4~\cite{OpenAI2023GPT4TR} have reached performance nearly on par with humans in various academic exams. The progress in LLMs has also inspired researchers to employ LLMs as an interface for multimodal tasks, such as vision-language learning~\cite{alayrac2022flamingo,li2023blip}, audio and speech recognition~\cite{gong2023listen,zhang2023speechgpt}, video understanding~\cite{2023videochat,chen2023videollm,zhang2023videollama}, \emph{etc}.

Among these tasks, vision-language learning is the most active field, with more than 50 vision LLMs proposed in the recent half-year alone~\cite{fu2023mme}. Typically, a vision LLM comprises a visual encoder, an LLM, and a projection module connecting the two components. The vision LLM is first trained on massive paired image-text data~\cite{schuhmann2022laion} for vision-language alignment and then fine-tuned on visual instruction datasets, enabling it to complete various instructions tied to visual inputs. Beyond vision, significant efforts have been invested in developing other modality-specific LLMs, such as audio~\cite{gong2023listen}, video~\cite{2023videochat}, and point clouds~\cite{guo2023pointbind}. These models generally mirror the architectural framework and training methodology of vision LLMs, and rely on the solid foundation of pretrained modality-specific encoders and well-curated instruction-tuning datasets for their effectiveness.

There are also several attempts to integrate multiple modalities into one MLLM~\cite{chen2023xllm,zhao2023chatbridge,han2023imagebind,moon2023anymal}. As an extension of vision LLM, most previous works align each modality with the LLM using modality-specific encoders and projection modules (middle of Fig.~\ref{fig:framework_small}). For instance, X-LLM~\cite{chen2023xllm} and ChatBridge~\cite{zhao2023chatbridge} connect pretrained image, video, and audio encoders with LLMs using separate Q-Former~\cite{li2023blip} or Perceiver~\cite{jaegle2021perceiver} models. However, these modality-specific encoders usually differ in architecture and considerable effort is required to unify them into a single framework. Furthermore, pretrained encoders that deliver reliable performance are usually restricted to widely used modalities such as image, audio, and video. This limitation poses a constraint on MLLMs' ability to expand to more modalities. Thus, a crucial challenge for MLLMs is \textit{how to build a unified and scalable encoder capable of handling a wide range of modalities.}

We get inspiration from recent works on transferring pretrained transformers to downstream modalities~\cite{lu2021pretrained,xu2022image2point,generalpatternmachines2023,zhang2023meta}. Lu \etal \cite{lu2021pretrained} proved that a frozen language-pretrained transformer can achieve strong performance on downstream modalities such as image classification. Meta-Transformer~\cite{zhang2023meta} demonstrated that a frozen visual encoder can achieve competitive results across 12 different data modalities. The insights from the works mentioned above suggest that pretrained encoders for each modality may not be necessary. Instead, a well-pretrained transformer may serve as a universal cross-modal encoder. 

In this paper, we present \textbf{OneLLM}, an MLLM that aligns eight modalities to language using one unified framework. As shown in Fig.~\ref{fig:framework_small}, OneLLM consists of lightweight modality tokenizers, a universal encoder, a universal projection module (UPM), and an LLM. In contrast to prior works, the encoder and projection module in OneLLM are shared across all modalities. The modality-specific tokenizers, each comprised of only one convolution layer, convert input signals into a sequence of tokens. Additionally, we add learnable modality tokens to enable modality switching and transform input tokens of diverse lengths into tokens of a fixed length.

Training a model of this complexity from scratch poses significant challenges. We start from a vision LLM and align other modalities to the LLM in a progressive way. Specifically, \textbf{(i)} we build a vision LLM with pretrained CLIP-ViT~\cite{radford2021learning} as the image encoder, accompanied by several transformer layers as the image projection module, and LLaMA2~\cite{touvron2023llama} as the LLM. After pretraining on massive paired image-text data, the projection module learns to map visual representations into the embedding space of LLM. \textbf{(ii)} To align with more modalities, we need a universal encoder and projection module. As discussed before, the pretrained CLIP-ViT is possible to serve as a universal encoder. For UPM, we propose to mix multiple image projection experts as a universal X-to-language interface. To increase the model capability, we also design a dynamic router to control the weight of each expert for the given inputs, which turns UPM into soft mixtures-of-experts~\cite{puigcerver2023sparse}. Finally, we progressively align more modalities with the LLM based on their data magnitude.

We also curate a large-scale multimodal instruction dataset, including captioning, question answering, and reasoning tasks across eight modalities: image, audio, video, point clouds, depth/normal map, Inertial Measurement Unit (IMU), and functional Magnetic Resonance Imaging (fMRI). By finetuning on this dataset, OneLLM has strong multimodal understanding, reasoning, and instruction-following capabilities. We evaluate OneLLM on multimodal captioning, question answering and reasoning benchmarks where it achieves superior performance than previous specialized models and MLLMs. In conclusion, we summary our contributions as:
\begin{itemize}
    \item We propose a unified framework to align multimodal inputs with language. Different from existing works with modality-specific encoders, we show that a unified multimodal encoder, which leverages a pretrained vision-language model and a mixture of projection experts, can serve as a general and scalable component for MLLMs.
    \item To the best of our knowledge, OneLLM is the first MLLM that integrates eight distinct modalities within a single model. With the unified framework and progressive multimodal alignment pipeline, OneLLM can be easily extended to incorporate more data modalities. 
    \item We curate a large-scale multimodal instruction dataset. OneLLM finetuned on this dataset achieves superior performance on multimodal tasks, outperforming both specialist models and existing MLLMs.
\end{itemize}

\begin{figure*}[t]
    \centering
    \includegraphics[width=\textwidth]{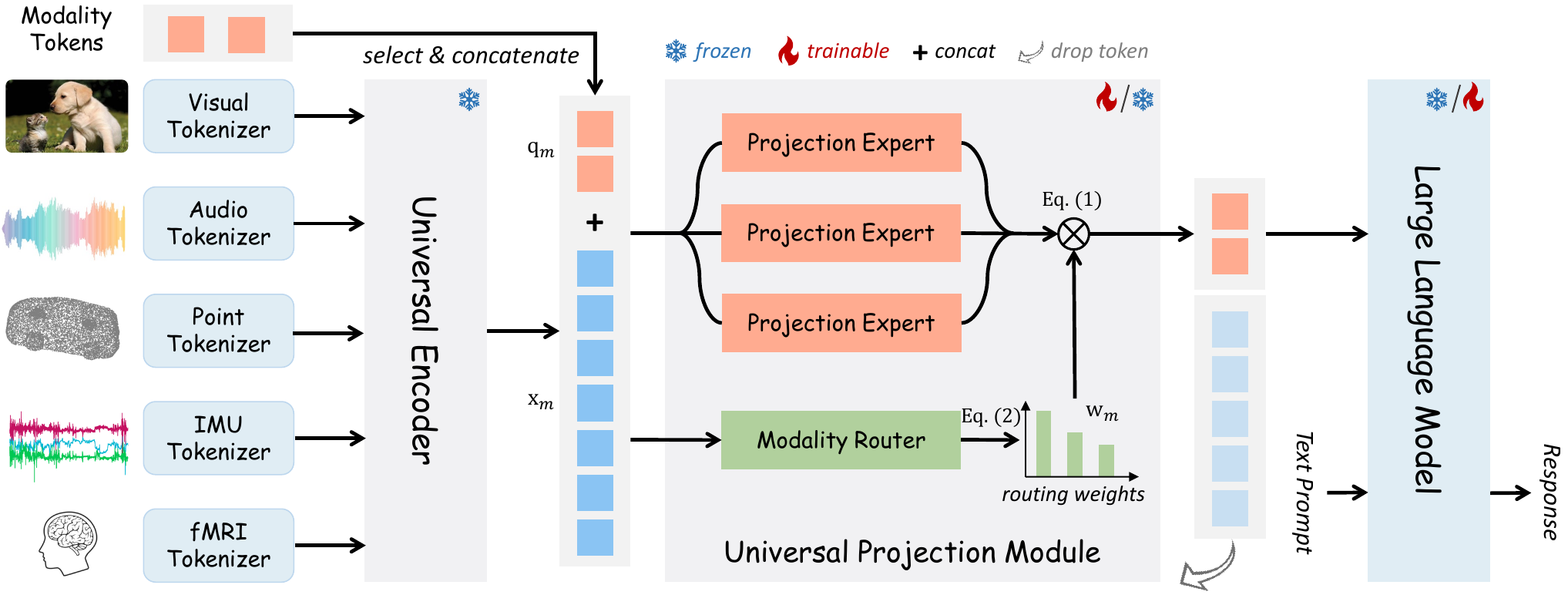}
    \vspace{-2mm}
    \caption{\textbf{The Architecture of OneLLM}. OneLLM consists of modality tokenizers, a universal encoder, a universal projection module (UPM) and an LLM. The modality tokenizer is a 2D/1D convolution layer to transform the input signal into a sequence of tokens. For simplicity, we omit video, depth/normal map tokenizers. The universal encoder is a frozen vision-language model (\emph{i.e.} CLIP~\cite{radford2021learning}) to extract high dimensional features. The UPM is composed of several projection experts and modality routers to align the input signal with language. For the alignment stage, we train modality tokenizers and UPM, and keep LLM frozen. For the instruction tuning stage, we only train the LLM and keep other models frozen. In a forward pass of UPM, we concatenate the input and modality tokens as input. Then we only take the modality tokens as a summary of the input signal and feed it into LLM for multimodal understanding.}
    \vspace{-1mm}
    \label{fig:framework_full}
\end{figure*}

\section{Related Work}

\noindent \textbf{Large Vision-Language Models.} Large Language Models (LLMs) have gained a lot of attention recently. Therefore, extending LLMs to the vision domain is an emergent and rapidly growing research area. Flamingo~\cite{alayrac2022flamingo} is a pioneer to inject frozen visual features into LLM with cross-attention layers, achieving superior performance on a wide range of vision-language tasks.
BLIP2~\cite{li2023blip} uses a Q-Former to aggregate visual features into a few tokens aligned with LLM. Recently, with the popularity of instruction-following LLMs, vision LLMs have experienced a new explosion. LLaMA-Adapter~\cite{gao2023llama,zhang2023llama} connects pretrained CLIP~\cite{radford2021learning} and LLaMA~\cite{touvron2023llama} with parameter-efficient fine-tuning methods, which can tackle close-set visual question answering and image captioning tasks. Subsequent works~\cite{liu2023improvedllava,zhu2023minigpt,gao2023llama,ye2023mplugowl} propose to train such model on large-scale image-text data, enabling it to complete various instructions about images. Among them, LLaVA~\cite{liu2023improvedllava} adopt a linear layer to directly project visual tokens into LLMs, while MiniGPT-4~\cite{zhu2023minigpt} and some other works~\cite{gao2023llama,ye2023mplugowl} resample visual tokens into fixed-length tokens, reducing the computation cost of LLMs. Our work also belongs to the later branch. We preset learnable tokens for each modality (\emph{i.e.}, modality tokens), which are then used to aggregate input information and generate fixed-length tokens for all modalities.

\noindent \textbf{Multimodal Large Language Models.} In addition to vision LLMs, recent works proposed to extend LLMs to other modalities, such as audio~\cite{gong2023listen,zhang2023speechgpt}, video~\cite{2023videochat,chen2023videollm,zhang2023videollama} and point cloud~\cite{xu2023pointllm,guo2023pointbind}. These works make it possible to unify multiple modalities into one LLM. X-LLM~\cite{chen2023xllm} adopts modality-specific Q-Former~\cite{li2023blip} and adapters to connect pretrained image, audio and video encoders with LLMs. ChatBridge~\cite{zhao2023chatbridge} and AnyMAL~\cite{moon2023anymal} follow a similar architecture with X-LLM but adopts Perceiver~\cite{jaegle2021perceiver} and linear layers respectively to align modality encoders with LLMs. Meanwhile, PandaGPT~\cite{su2023pandagpt} and ImageBind-LLM~\cite{han2023imagebind} utilize ImageBind~\cite{girdhar2023imagebind} as the modality encoder and therefore naturally support multimodal inputs. However, current MLLMs are limited to supporting common modalities such as image, audio and video. It remains unclear how to expand MLLMs to more modalities with a unified framework. In this work, we propose a unified multimodal encoder to align all modalities with language. We show that one universal encoder and projection module can effectively map multimodal inputs to LLM. To our knowledge, OneLLM is first MLLM capable of supporting eight distinct modalities.

\noindent \textbf{Multimodal-Text Alignment.} Aligning multiple modalities into one joint embedding space is important for cross-modal tasks, which can be divided into two lines of works: discriminative alignment and generative alignment. The most representative work of discriminative alignment is CLIP~\cite{radford2021learning}, which utilize contrastive learning to align image and text. Follow-up works extend CLIP to audio-text~\cite{guzhov2022audioclip,wu2023clap}, video-text~\cite{xu2021videoclip,luo2022clip4clip}, point-text~\cite{zhang2022pointclip} \emph{etc.} Besides, ImageBind~\cite{girdhar2023imagebind} proposes to bind various modalities to images with contrastive learning. On the other hand, generative alignment has attracted much attention in the era of LLM. GIT~\cite{wang2022git} aligns image and text using a generative image-to-text transformer. BLIP2~\cite{li2023blip} proposes generative pretraining to connect frozen vision encoder and LLM. VALOR~\cite{chen2023valor} and VAST~\cite{chen2023vast} extends the training paradigm of BLIP2 to more modalities such as audio and video. Our work also belongs to generative alignment. In contrast to prior works, we directly align mutlimodal inputs to LLMs, thus getting rid of the stage of training modality encoders.

\section{Method}

In this section, we will first introduce the architecture of OneLLM (Sec.~\ref{subsec:architecture}) and then present our two training phases: progressive multimodal alignment (Sec.~\ref{subsec:alignment}) and unified multimodal instruction tuning (Sec.~\ref{subsec:instruction_tuning}).

\subsection{Model Architecture}
\label{subsec:architecture}

Fig.~\ref{fig:framework_full} depicts the four main components of OneLLM: modality-specific tokenizers, a universal encoder, a universal projection module (UPM) and an LLM. Detailed descriptions are presented in the following sections.

\noindent \textbf{Lightweight Modality Tokenizers.} The modality tokenizer is to transform the input signal into a sequence of tokens, thereby a transformer-based encoder can process these tokens. We denote the input tokens as $\mathbf{x} \in \mathbb{R}^{L \times D}$, where $L$ is the sequence length and $D$ is the token dimension.  Considering the variations inherent to different data modalities, we design a separate tokenizer for each modality. For visual inputs with 2D position information such as image and video, we directly utilize a single 2D convolution layer as the tokenizer. For other modalities, we transform the input into a 2D or 1D sequence, which is then tokenized using a 2D/1D convolution layer. For example, we transform audio signals into 2D spectrogram and sample a subset of point clouds with 2D geometric prior. Due to space limit, please refer to Sec.~\ref{subsec:appendix_tokenizer} of the appendix for more details.

\noindent \textbf{Universal Encoder.} As discussed in Sec.~\ref{sec:intro}, frozen pretrained transformers demonstrate strong modality transfer capability~\cite{lu2021pretrained,zhang2023meta}. Therefore, we leverage pretrained vision-language models as the universal encoder for all modalities. Vision-language models, when trained on extensive image-text data, typically learn robust alignment between vision and language, so they can be easily transferred to other modalities. In OneLLM, we use CLIP-ViT~\cite{radford2021learning} as a universal computation engine. Following previous works~\cite{lu2021pretrained,zhang2023meta}, we keep the parameters of CLIP-ViT frozen during training. Note that for video signals, we will feed all video frames into the encoder in parallel and perform token-wise averaging between frames to speed up training. Other strategies, such as token concatenation, may further enhance the model's video understanding capability.

\noindent \textbf{Universal Projection Module.} In contrast to existing works with modality-specific projection, we propose a Universal Projection Module (UPM) to project any modality into LLM's embedding space. As shown in Fig.~\ref{fig:framework_full}, UPM consists of $K$ projection experts $\{P_k\}$, where each expert is a stack of transformer layers pretrained on image-text data (will discuss in Sec.~\ref{subsec:alignment}). Although one expert can also realize any modality-to-LLM projection, our empirical findings suggest that multiple experts are more effective and scalable. When scaling to more modalities, we only need to add a few parallel experts.

To integrate multiple experts into one module, we propose a dynamic \textit{modality router} $R$ to control each expert's contribution and increase the model capacity. The router $R$ is structured as a straightforward Multi-Layer Perception that receives input tokens and calculates the routing weights for each expert, \emph{i.e.}, a soft router~\cite{puigcerver2023sparse}. We will also discuss other types of router in Sec.~\ref{subsec:ablation}, such as constant router and sparse router.
Besides, we add learnable modality tokens $\{\mathbf{q}_m\}_{m \in \mathcal{M}}$ to switch between modalities, where $\mathcal{M}$ is the set of modalities and $\mathbf{q}_m \in \mathbb{R}^{N\times D}$ contains $N$ tokens of dimension $D$. In a forward pass for modality $m$, we feed the concatenation of input tokens $\mathbf{x}_m \in \mathbb{R}^{L \times D}$ and modality tokens $\mathbf{q}_m$ into UPM:{\small
\begin{align}
    [\mathbf{\bar q}_m, \mathbf{\bar x}_m] &= \mathrm{UPM}([\mathbf{q}_m, \mathbf{x}_m]) = \sum_{k=1}^K \mathbf{w}_m \cdot P_k([\mathbf{q}_m, \mathbf{x}_m]), \\
    \mathbf{w}_m &= \sigma \circ R_m([\mathbf{q}_m, \mathbf{x}_m]),
\end{align}
}where $\mathbf{w}_m \in \mathbb{R}^{N\times K}$ is the routing weight and the SoftMax function $\sigma$ is to ensure $\sum_{k=1}^K \mathbf{w}_{m,k}=1$. For any modality $m$, we \textit{only} extract the projected modality tokens $\mathbf{\bar q}_m$ as a summary of input signals, transforming $\mathbf{x}_m$ from varying lengths into uniform, fixed-length tokens.

\noindent \textbf{LLM.} We employ the open-source LLaMA2~\cite{touvron2023llama2} as the LLM in our framework. The input to LLM includes projected modality tokens $\mathbf{\bar q}_m$ and the text prompt after word embedding. Note we always put modality tokens at the beginning of the input sequence for simplicity. Then LLM is asked to generate appropriate response conditioned on modality tokens and text prompt.

\subsection{Progressive Multimodal Alignment}
\label{subsec:alignment}

Image-text alignment has been well investigated in previous works~\cite{llava1,zhu2023minigpt,gao2023llama}. Therefore, a naive approach for multimodal alignment is to jointly train the model on multimodal-text data. However, training models directly on multimodal data can lead to biased representations between modalities due to the imbalance of data scale. Here we propose to train an image-to-text model as initialization and progressively ground other modalities into LLM.

\noindent \textbf{Image-Text Alignment.} We begin with a basic vision LLM framework, comprising an image tokenizer, a pretrained CLIP-ViT, an image projection module $P_I$ and an LLM. Considering that image-text data is relatively abundant compared to other modalities, we first train the model on image-text data to well align CLIP-ViT and LLM, \emph{i.e.}, learning a good image-to-text projection module. The pretrained $P_I$ not only serves as a bridge connecting images and language, but also provides a good initialization for multimodal-text alignment.
Then we build $\mathrm{UPM}$ by mixing multiple pretrained $P_I$: $ \mathrm{UPM} = \{ P_k \} = \{ \mathrm{Init}(P_I) \}$, where $\mathrm{Init}$ is weight initialization, which effectively reduces the cost of aligning other modalities to language.

\noindent \textbf{Multimodal-Text Alignment.} We formulate multimodal-text alignment as a continual learning process~\cite{van2019three}. At timestamp $t$, we have trained the model on a set of modalities $\mathcal{M}_1 \cup \mathcal{M}_2 \cdots \mathcal{M}_{t-1}$, and the current training data is from $\mathcal{M}_t$. To prevent catastrophic forgetting, we will sample evenly from both previous trained data and current data. In our case, we divide multimodal-text alignment into multiple training stages based on their data magnitude: stage I (image), stage II (video, audio and point cloud) and stage III (depth/normal map, IMU and fMRI). If we want to support new modalities, we can repeat the training episode, \emph{i.e.}, sampling a similar amount of data from previous modalities and jointly training the model with the current modalities.

\noindent \textbf{Multimodal-Text Dataset.} We collect X-text pairs for each modality. The image-text pairs include LAION-400M~\cite{schuhmann2022laion} and LAION-COCO~\cite{laion_coco}. The training data for video, audio and point clouds are WebVid-2.5M~\cite{webvid}, WavCaps~\cite{mei2023wavcaps} and Cap3D~\cite{cap3d}, respectively. Since there is no large-scale depth/normal map-text data, we use pretrained DPT model~\cite{dpt,eftekhar2021omnidata} to generate depth/normal map. The source images and text and from CC3M~\cite{sharma2018conceptual}. For IMU-text pairs, we use the IMU sensor data of Ego4D~\cite{grauman2022ego4d}. For fMRI-text pairs, we use fMRI signals from the NSD~\cite{nsd} dataset and take the captions associated with the visual stimuli as text annotations. Note that the input to LLM is the concatenation of modality tokens and caption tokens. We do not add system prompts at this stage to reduce the number of tokens and speed up training.

\subsection{Unified Multimodal Instruction Tuning}
\label{subsec:instruction_tuning}

After multimodal-text alignment, OneLLM becomes a multimodal captioning model which can generate a short description for any input. To fully unleash OneLLM's multimodal understanding and reasoning capabilities, we curate a large-scale multimodal instruction tuning dataset to further finetune OneLLM.

\noindent \textbf{Multimodal Instruction Tuning Dataset.} We collect instruction tuning (IT) dataset for each modality. Following previous works~\cite{instructblip,liu2023improvedllava}, the \textit{image} IT datasets are sampled from the following datasets: LLaVA-150K~\cite{llava1}, COCO Caption~\cite{cococap}, VQAv2~\cite{goyal2017vqav2}, GQA~\cite{hudson2019gqa}, OKVQA~\cite{okvqa}, A-OKVQA~\cite{schwenk2022okvqa}, OCRVQA~\cite{mishra2019ocrvqa}, RefCOCO~\cite{kazemzadeh2014referitgame} and Visual Genome~\cite{krishna2017visual}. The \textit{video} IT datasets include MSRVTT-Cap~\cite{xu2016msrvtt}, MSRVTT-QA~\cite{msvd} and video instruction data from~\cite{zhao2023chatbridge}. The \textit{audio} IT datasets include AudioCaps~\cite{kim2019audiocaps} and audio conversation data from~\cite{zhao2023chatbridge}. The \textit{point cloud} IT dataset is a 70K point cloud description, conversation and reasoning dataset from~\cite{xu2023pointllm}. The \textit{depth/normal map} IT datasets are generated from image IT datasets: we random sample 50K visual instruction data from LLaVA-150K and generate depth/normal map using DPT model~\cite{eftekhar2021omnidata}. For \textit{IMU} and \textit{fMRI} IT datasets, we also random sample a subset from Ego4D~\cite{grauman2022ego4d} and NSD~\cite{nsd}, respectively. Finally, our mutlimodal IT datasets have about \textbf{2M} items, covering multiple tasks such as detailed description/reasoning, conversation, short question answering and captioning.

\noindent \textbf{Prompt Design.} Given the diverse modalities and tasks within our multimodal IT datasets, we carefully design the prompts to avoid conflicts between them. \textbf{(a)} When utilizing IT datasets generated by GPT4 (\emph{e.g.}, LLaVA-150K), we adopt the original prompts provided by these datasets. \textbf{(b)} For captioning tasks, we empoly the prompt: \textit{Provide a one-sentence caption for the provided \{modal\}}. \textbf{(c)} For open-ended question answering tasks, we enhance the question with \textit{Answer the question using a single word or phrase}. \textbf{(d)} For question answering tasks with options, the prompt is: \textit{\{Question\} \{Options\} Answer with the option's letter from the given choices directly}. \textbf{(e)} For IMU and fMRI datasets, we apply prompt such as \textit{Describe the motion} and \textit{Describe this scene based on fMRI data}. Despite using these fixed prompts, our experiments indicate that OneLLM is capable of generalizing to open-ended prompts during inference. For detailed prompts on each task and modality, please check out Sec.~\ref{subsec:prompt_design} of the appendix.

In the instruction tuning stage, we organize the input sequence as: $\{{\bar q}, Sys, [Ins_t, Ans_t]_{t=1}^T\}$ where ${\bar q}$ is the modality tokens, $Sys$ is the system prompt, $[Ins_t, Ans_t]$ corresponds to the $t$-th instruction-answer pair in a conversation. Note that for multimodal inputs involving multiple modalities, such as audio-visual tasks~\cite{li2022musicvavqa}, we position all modality tokens at the start of the input sequence.

We fully finetune the LLM and keep rest parameters frozen. Although recent works often employ parameter-efficient methods~\cite{hu2021lora}, we empirically show that the full finetuning approach more effectively harnesses the multimodal capabilities of OneLLM, particularly with the utilization of smaller LLMs (\emph{e.g.}, LLaMA2-7B).

\input{tables/image_eval}

\input{tables/video_eval}

\input{tables/audio_eval}

\input{tables/audio_visual_eval}
\input{tables/point_eval}
\input{tables/depth_normal_eval}

\section{Experiment}

\subsection{Implementation Details}
\noindent \textbf{Architecture.} The universal encoder is CLIP VIT Large pretrained on LAION~\cite{schuhmann2022laion}. The LLM is LLaMA2-7B~\cite{touvron2023llama2}. The UPM has $K$=3 projection experts, where each expert has eight Transformer blocks and 88M parameters. The size of modality tokens for each modality is $\mathbb{R}^{30 \times 1024}$. 

\noindent \textbf{Training Details.} We use AdamW optimizer with $\beta_1$=0.9, $\beta_2$=0.95 and weight decay of 0.1. We apply a linear learning rate warmup during the first 2K iterations. For stage I, we train OneLLM on 16 A100 GPUs for 200K iterations. The effective batch size (using gradient accumulation) is 5120. The maximum learning rate is 5e-5. For stage II (\textit{resp.} III), we train OneLLM on 8 GPUs for 200K (\textit{resp.} 100K) with an effective batch size of 1080 and maximum learning rate of 1e-5. In the instruction tuning stage, we train OneLLM on 8 GPUs for 1 epoch (96K) with an effective batch size of 512 and maximum learning rate of 2e-5.

\subsection{Quantitative Evaluation}

We evaluate OneLLM on multimodal tasks and put evaluation details to Sec.~\ref{sec:eval_details} of the appendix.

\noindent \textbf{Image-Text Evaluation.} In Tab.~\ref{tab:image_eval}, we evaluate OneLLM on visual question answering (VQA), image captioning and recent multimodal benchmarks. For VQA tasks, OneLLM-7B outperforms other MMLLMs such as ChatBridge-13B~\cite{zhao2023chatbridge} and AnyMAL-13B~\cite{moon2023anymal} by a large margin. Our 7B model is even better than AnyMAL with 70B parameters. For image captioning tasks, OneLLM-7B is on-par with ChatBridge-13B. Although OneLLM is not specifically designed for vision tasks, our results demonstrate that OneLLM can also reach the leading level in vision specialized LLMs, and the gap between MMLLMs and vision LLMs has further narrowed.

\noindent \textbf{Video-Text Evaluation.} As shown in Tab.~\ref{tab:video_eval}, we evaluate OneLLM on video QA and captioning tasks. Our model outperforms both MLLMs (ChatBridge and AnyMAL) and video-specific models (FrozenBiLM~\cite{yang2022zero} and InternVideo~\cite{wang2022internvideo}) in video QA tasks. Notably, our training datasets do not include video QA data like NextQA~\cite{nextqa} and How2QA~\cite{li2020hero}, which are video QA tasks that provide answer options. 
However, our model's training on similar VQA datasets (\emph{e.g.}, A-OKVQA~\cite{schwenk2022okvqa}) has evidently enhanced its emergent cross-modal capabilities, contributing to the improved performance in video QA tasks.

\noindent \textbf{Audio-Text Evaluation.} We evaluate OnLLM on audio captioning and QA tasks. In Tab.~\ref{tab:audio_eval}, we outperforms both ChatBridge and LTU~\cite{gong2023listen} on Clotho Caption~\cite{drossos2020clotho}. Notably, our zero-shot result on Clotho AQA~\cite{lipping2022clotho} is on par with fully finetuned Pengi~\cite{deshmukh2023pengi}. Similar to our conclusion on video QA, we believe that the captioning task requires more dataset-specific training, while the QA task may be a more accurate measure of the model's inherent zero-shot understanding capabilities.

\noindent \textbf{Audio-Video-Text Evaluation.} We evaluate OneLLM on audio-video-text tasks, such as QA (MUSIC AVQA~\cite{li2022musicvavqa}), captioning (VALOR-32K~\cite{chen2023valor}) and dialog completion (AVSD~\cite{alamri2019audiovisual}) based on the video and background audio. As shown in Tab.~\ref{tab:audio_visual_eval}, OneLLM-7B surpasses ChatBridge-13B on all three datasets. Note that ChatBridge was trained on an audio-visual dataset~\cite{chen2023valor}, while OneLLM has not been trained on any audio-visual datasets. Since all modalities in OneLLM are well aligned with language, we can directly input video and audio signals to OneLLM during inference. 

\noindent \textbf{Point Cloud-Text Evaluation.} In Tab.~\ref{tab:point_eval}, We evaluate OneLLM on point cloud captioning and classification tasks. OneLLM can achieve excellent captioning results due to our carefully designed instruction prompts for switching between tasks (Sec.~\ref{subsec:instruction_tuning}), while InstructBLIP~\cite{instructblip} and PointLLM~\cite{xu2023pointllm} struggle to generate short and accurate captions. On the classification task, OneLLM can also achieve comparable results to PointLLM.

\noindent \textbf{Depth/Normal Map-Text Evaluation.} Since there are currently no QA and captioning tasks using depth/normal maps, we evaluate OneLLM on two scene classification datasets~\cite{nyuv2,song2015sun}. The performance, as displayed in Tab.~\ref{tab:depth_normal_eval}, reveals that OneLLM achieves superior zero-shot classification accuracy compared to CLIP. These results affirm that OneLLM trained on synthetic depth/normal map data can adapt to real world scenarios.

\noindent \textbf{IMU-Text and fMRI-Text Evaluation.} Since IMU/fMRI to text generation are seldom explored in previous literature, we solely report our results on IMU/fMRI captioning. For IMU captioning on Ego4D~\cite{grauman2022ego4d}, we evaluate OneLLM on a held-out subset with 2000 items. The CIDEr and ROUGE-L score are 24.9 and 19.5, respectively. For fMRI captioning on NSD~\cite{nsd}, we evaluate OneLLM on its testing set, where OneLLM achieves 31.7 CIDEr and 25.1 ROUGE-L.

\input{tables/ablation}
\subsection{Ablation Experiments}
\label{subsec:ablation}

\begin{figure*}[t]
    \centering
    \includegraphics[width=\textwidth]{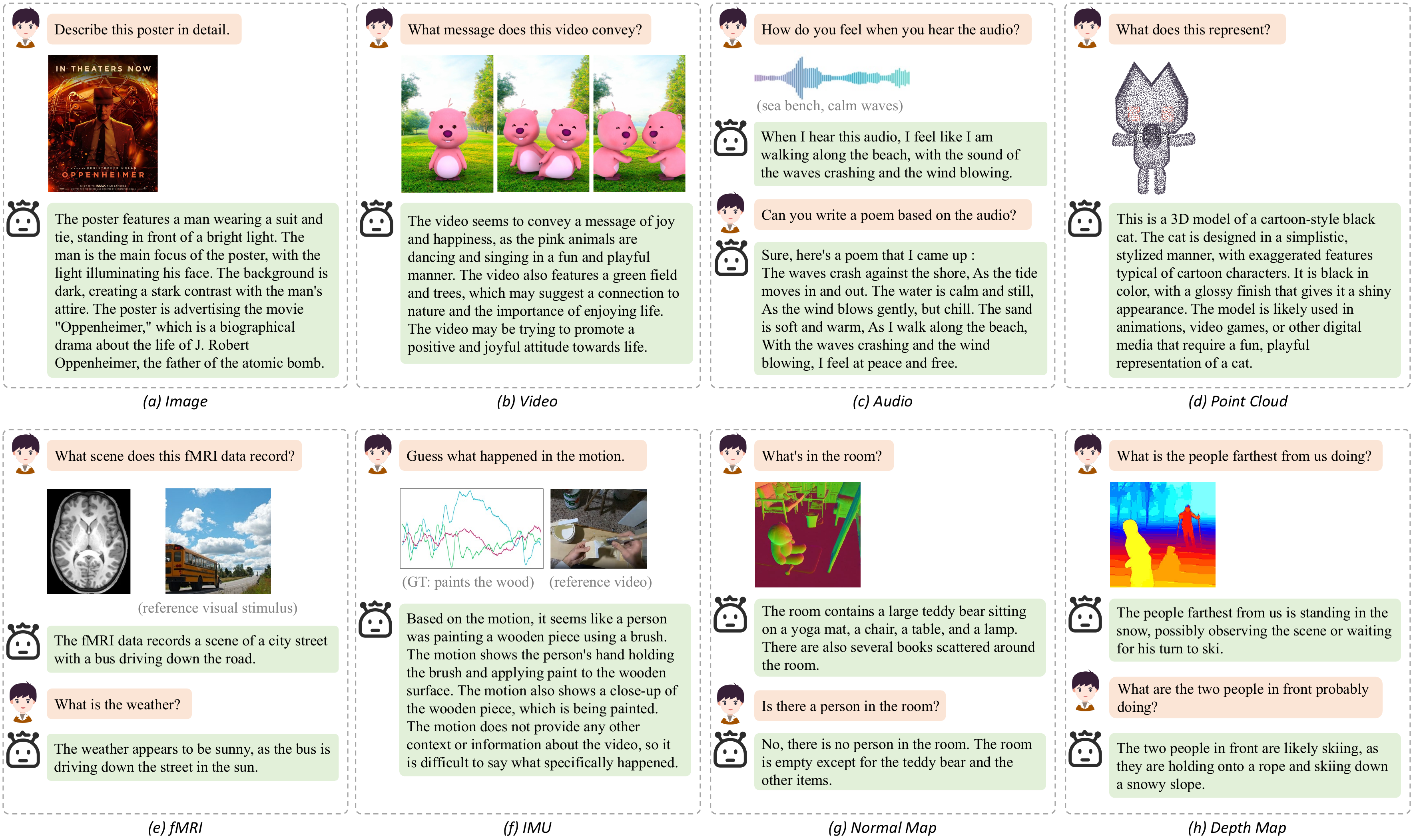}
    \vspace{-5mm}
    \caption{\textbf{Qualitative Results on Eight Modalities.} All demo inputs are from the web or the testing set of corresponding modalities.}
    \vspace{-3mm}
    \label{fig:demos}
\end{figure*}

In this section, we will explore some key designs of OneLLM. Our ablation experiments are conducted on a subset of the training data, which only includes multimodal alignment and instruction tuning datasets of image, audio and video, except for studies on the number of experts. Other settings remain unchanged if not specified.

\noindent \textbf{Separate Training \emph{vs.} Joint Training.} An important question for MLLMs is whether a jointly trained MLLM is better than modality-specific MLLM? To address this, we compare the performance of separately trained MLLMs against a jointly trained MLLM in Tab.~\ref{tab:ablation} (a). In separate training, the model can only access its own data; in joint training, the model is jointly trained on all data. On two image-text tasks NoCaps and VQAv2, we can see that separately and jointly trained models achieve comparable results; While separately trained audio and video models are much worse than the jointly trained model on ClothoQA and MSVDQA, respectively. This suggest that joint training substantially benefits data-scarce modalities (\emph{e.g.}, audio and video), by allowing for the transfer of learned knowledge (\emph{e.g.}, question answering) across modalities.

\noindent \textbf{Image Alignment Benefits Multimodal Alignment.} Tab.~\ref{tab:ablation} (b) demonstrate that OneLLM with image-text alignment can help multimodal-text alignment. If we directly align all modalities with text using a random initialized model (\emph{i.e.} universal projection module), the performance on image and video will drop significantly. Instead, OneLLM with image-text pretraining can better balance different modalities.

\noindent \textbf{Number of Projection Experts.} The number of projection experts in UPM is closely related to the number of modalities that OneLLM can accommodate. As shown in Tab.~\ref{tab:ablation}, OneLLM with three projection experts is enough to hold all modalities. Increasing the number of experts does not bring about the desired improvement, while the results with one expert is also not satisfactory.

\noindent \textbf{Router Type.} The modality router is to link multiple projection experts into a single module. Here we discuss three types of router: constant router, sparse router and the default soft router. \textbf{(a)} Constant router links $K$ experts with a constant number $1/K$. The output of constant router is $\sum^K_{k=1} \frac{1}{K} \cdot P_k (\mathbf{x})$. \textbf{(b)} Sparse router only selects one expert with the maximum routing weight. The output is $w_{k^*} P_{k^*}(\mathbf{x})$ where $k^*=\mathop{\arg\max}\limits_{k} w_k$. As shown in Tab.~\ref{tab:ablation} (d), soft router outperforms other two routers, indicating its effectiveness for dynamic routing of multimodal signals.

\subsection{Qualitative Analysis}

Fig.~\ref{fig:demos} gives some qualitative results of OneLLM on eight modalities. We show OneLLM can \textbf{(a)} understand both visual and textual content in images, \textbf{(b)} leverage temporal information in videos, \textbf{(c)} do creative writing based on audio content, \textbf{(d)} understand the details of 3D shapes, \textbf{(e)} analyze visual scenes recorded in fMRI data, \textbf{(f)} guess the person's action based on motion data, and \textbf{(g)-(h)} scene understanding using depth/normal map. Due to space limit, we put more qualitative results to Sec.~\ref{sec:qualitative} of the appendix.

\section{Conclusion}

In this work, we introduce OneLLM, an MLLM that aligns eight modalities with language using a unified framework. Initially, we first train a basic vision LLM. Building on this, we design a multimodal framework with a universal encoder, a UPM and an LLM. By a progressive alignment pipeline, OneLLM can handle multimodal inputs with a single model. Furthermore, we curate a large-scale multimodal instruction dataset to fully unleash OneLLM's instruction-following capability. Finally, we evaluate OneLLM on 25 diverse benchmarks, showing its excellent performance.

\noindent \textbf{Limitation and Future Work.} Our work faces two primary challenges: \textbf{(i)} The absence of large-scale, high-quality datasets for modalities beyond image, which leads to a certain gap between OneLLM and specialized models on these modalities. \textbf{(ii)} Fine-grained multimodal understanding in high-resolution images, long sequences video and audio \emph{etc.} In the future, we will collect high-quality datasets and design new encoders to realize fine-grained multimodal understanding, \emph{e.g.}, supporting varying length inputs~\cite{fuyu-8b}.

\noindent{\textbf{Acknowledgements.}} {\small This work is partially supported by the National Natural
Science Foundation of China (Grant No. 62306261), CUHK Direct Grants (Grant No. 4055190), National Key R\&D Program of China (2022ZD0160201) and Shanghai Artificial Intelligence Laboratory.}

\appendix

\section{Appendix Overview}
\begin{itemize}
    \item Sec.~\ref{sec:ablation}: Additional Ablation Experiments.
    \item Sec.~\ref{sec:impl}: Additional Implementation Details.
    \item Sec.~\ref{sec:eval_details}: Evaluation Details.
    \item Sec.~\ref{sec:comp}: Comparison with Prior Works.
    \item Sec.~\ref{sec:qualitative}: Additional Qualitative Results.
\end{itemize}

\section{Additional Ablation Experiments}
\label{sec:ablation}

\begin{table}[h]
    \centering
    \resizebox{\linewidth}{!}{%
    \begin{tabular}{cc|c|llll}
    \toprule
    \multicolumn{2}{c|}{Encoder} & \multirow{2}{*}{Mem.} & \multirow{2}{*}{Nocaps} & \multirow{2}{*}{VQAv2} & \multirow{2}{*}{\makecell[c]{ClothoQA}} & \multirow{2}{*}{\makecell[c]{MSVDQA}} \\
    Type        & Frozen         & &                         &                        &                           &                         \\
    \midrule
    \rowcolor{Gray}
    CLIP        & \ding{51}     & 46Gb &  115.8                  & 71.6                   & 57.4                      & 56.8                        \\
    CLIP        & \ding{55}     & 74Gb &  106.0{\footnotesize(\textcolor{Red}{-9.8})}                  &  69.1{\footnotesize(\textcolor{Red}{-2.5})}                  & 62.1{\footnotesize(\textcolor{Green}{+4.7})}                      & 53.6{\footnotesize(\textcolor{Red}{-3.2})}                       \\
    DINOv2      & \ding{51}     & 33Gb &  104.6{\footnotesize(\textcolor{Red}{-11.2})}                  &  67.0{\footnotesize(\textcolor{Red}{-4.6})}                  &    56.8{\footnotesize(\textcolor{Red}{-0.6})}                       & 54.7{\footnotesize(\textcolor{Red}{-2.1})}                       \\
    \bottomrule
    \end{tabular}%
    }
    \vspace{-2mm}
    \caption{\textbf{Ablation Experiments on Universal Encoder.}}
    \label{tab:appendix_ablation}
    \end{table}

In the main paper, we follow previous works~\cite{zhang2023meta} and set a frozen CLIP-ViT as the universal encoder. Here we explore other design choices such as trainable CLIP-ViT and DINOv2~\cite{oquab2023dinov2} as the encoder.


\paragraph{Frozen \emph{vs.} Trainable Encoder.} We first turn on all the parameters in the multimodal-text alignment stage. As shown in Tab.~\ref{tab:appendix_ablation}, the performance for visual modalities (image and video) dropped significantly, while the result for audio QA (ClothoQA) improved by 4.7\%. We think trainable CLIP will break the pretrained vision-language representations but can leave more space for learning other modalities. However, considering the memory usage (46Gb \emph{vs.} 74Gb), frozen CLIP will be a better choice for our framework.

\paragraph{Beyond Vision-Language Encoder.} In addition to the vision-language encoder CLIP-ViT, we also explore other models, such as the self-supervised vision model DINOv2~\cite{oquab2023dinov2}, as the universal encoder. In Tab.~\ref{tab:appendix_ablation}, we noticed that the performance of OneLLM using DINOv2 is lower than the model using CLIP-ViT because DINOv2 is not aligned with language and we need to learn the vision-language alignment from scratch.

\vspace{-1mm}
\section{Additional Implementation Details}
\label{sec:impl}
\subsection{Lightweight Modality Tokenizers}
\label{subsec:appendix_tokenizer}

The modality tokenizer is to transform input signal into a sequence of tokens. Here we will introduce the tokenizer of each modality in detail.

\paragraph{Visual Tokenizer.} We use the same tokenizer setting for visual modalities, \emph{i.e.}, image, video, depth/normal map. The visual tokenizer is a single 2D convolution layer:{\footnotesize
\begin{align}
    \mathrm{Conv2D}(C_{in}=3, C_{out}=1024, K=(14, 14), S=(14, 14)),
\end{align}
}where $C_{in}$, $C_{out}$, $K$ and $S$ denote the input channel, output channel, kernel size and stride, respectively. Note that for a video input $\mathbf{x} \in \mathbb{R}^{T \times H \times W}$ with $T$ frames, height $H$ and width $W$, we parallel feed its frames into the tokenizer, resulting in $T\times \frac{H}{14} \times \frac{W}{14}$ tokens. Similarly, image, depth/normal map can also be regarded as a one-frame video input $\mathbf{x} \in \mathbb{R}^{1 \times H \times W}$.

\paragraph{Audio Tokenizer.} We first transform audio signals into 2D spectrogram features $\mathbf{x} \in \mathbb{R}^{1 \times H \times W}$, where $H$=128 and $W$=1024 by default. Following~\cite{gong21b_ast}, the audio tokenzier is a single 2D convolution layer:{\footnotesize
\begin{align}
    \mathrm{Conv2D}(C_{in}=1, C_{out}=1024, K=(16, 16), S=(10, 10)).
\end{align}
}

\paragraph{Point Tokenizer.} For a raw point cloud, we sample 8192 points using Furthest Point Sampling (FPS), resulting in a 2D tensor $\mathbf{x} \in \mathbb{R}^{8192 \times 6}$. Then we use the KNN algorithm to group these points into 512 groups: $\mathbf{x} \in \mathbb{R}^{512 \times 32 \times 6}$ where 32 is the size of each group. After that, we encode the point cloud with a 2D convolution layer:{\footnotesize
\begin{align}
    \mathrm{Conv2D}(C_{in}=6, C_{out}=1024, K=(1, 1), S=(1, 1)),
\end{align}
}followed by a \texttt{max} operation on dimension 1. Finally, the shape of output tokens is $\mathbb{R}^{1024 \times 1024}$.

\paragraph{IMU Tokenizer.} For an IMU input with shape $\mathbb{R}^{2000 \times 6}$, we tokenize it with a 1D convolution layer:
{\footnotesize
\begin{align}
    \mathrm{Conv1D}(C_{in}=6, C_{out}=1024, K=10, S=1),
\end{align}
}resulting in a sequence of tokens $\mathbf{x} \in \mathbb{R}^{1024 \times 391}$.

\paragraph{fMRI Tokenizer.} The shape of an fMRI signal is $\mathbb{R}^{15724}$. We tokenize it with a 1D convolution layer:
{\footnotesize
\begin{align}
    \mathrm{Conv1D}(C_{in}=15724, C_{out}=8196, K=1, S=1).
\end{align}
}We then resize the output tensor $\mathbf{x} \in \mathbb{R}^{8196}$ into a 2D tensor $\mathbf{x} \in \mathbb{R}^{1024 \times 8}$ to align with the input of the transformer encoder.

\subsection{Multimodal-Text Alignment Dataset}

We summary the multimodal-text alignment dataset in Tab.~\ref{tab:appendix_data}. For depth/normal-text pairs, we adopt DPT model~\cite{dpt} pretrained on ominidata~\cite{eftekhar2021omnidata} to generate depth/normal map. The source dataset is a subset of CC3M~\cite{sharma2018conceptual}, around 0.5M image-text pairs. For IMU-text pairs, we use the IMU sensor data of Ego4D~\cite{grauman2022ego4d} and the corresponding video narrations (\emph{i.e.}, text annotations). For fMRI-text pairs, we use the \texttt{subj01} imaging session of NSD~\cite{nsd} and follow the same data split with~\cite{scotti2023reconstructing}. Note that the visual stimulus, \emph{i.e.}, images shown to participants, are from MS COCO~\cite{cococap}. Therefore, we use the image captions in COCO Captions as text annotations of fMRI-text pairs.

\begin{table}[ht]
\centering
\small
\resizebox{\linewidth}{!}{%
\begin{tabular}{lrlrl}
\toprule
\multirow{2}{*}{Modality} & \multicolumn{2}{l}{Multimodal-Text Alignment}                                              & \multicolumn{2}{l}{Multimodal Instruction Tuning}                                                                                                                                                                                                                                                                  \\
                          & Size  & Dataset                                                                            & Size           & Dataset                                                                                                                                                                                                                                                                                           \\
\midrule
Image                     & 1000M & \makecell[l]{LAION-400M~\cite{schuhmann2022laion} \\ LAION-COCO~\cite{laion_coco}} & 1216K          & \makecell[l]{LLaVA-150K~\cite{llava1}\\COCO Caption~\cite{cococap}\\VQAv2~\cite{goyal2017vqav2}, GQA~\cite{hudson2019gqa}\\OKVQA~\cite{okvqa}, A-OKVQA~\cite{schwenk2022okvqa}\\ OCRVQA~\cite{mishra2019ocrvqa}, RefCOCO~\cite{kazemzadeh2014referitgame}\\Visual Genome~\cite{krishna2017visual}} \\
\midrule
Video                     & 2.5M  & WebVid-2.5M~\cite{webvid}                                                          & 461K           & \makecell[l]{MSRVTT-Cap~\cite{xu2016msrvtt}\\MSRVTT-QA~\cite{msvd}\\Video Conversation~\cite{zhao2023chatbridge}}                                                                                                                                                                                 \\
\midrule
Audio                     & 0.4M  & WavCaps~\cite{mei2023wavcaps}                                                      & 60K            & \makecell[l]{AudioCaps~\cite{kim2019audiocaps}\\Audio Conversation~\cite{zhao2023chatbridge}}                                                                                                                                                                                                                   \\
\midrule
Point                     & 0.6M  & Cap3D~\cite{cap3d}                                                                 & 70K            & Point Conversation~\cite{xu2023pointllm}                                                                                                                                                                                                                                                          \\
\midrule
Depth                     & 0.5M  & CC3M~\cite{sharma2018conceptual}                                                   & 50K            & LLaVA-150K~\cite{llava1}                                                                                                                                                                                                                                                                          \\
\midrule
Normal                    & 0.5M  & CC3M~\cite{sharma2018conceptual}                                                   & 50K            & LLaVA-150K~\cite{llava1}                                                                                                                                                                                                                                                                          \\
\midrule
IMU                       & 0.5M  & Ego4D~\cite{grauman2022ego4d}                                                      & 50K            & Ego4D~\cite{grauman2022ego4d}                                                                                                                                                                                                                                                                     \\
\midrule
fMRI                      & 9K    & NSD~\cite{nsd}                                                                     & 9K             & NSD~\cite{nsd}                                                                                                                                                                                                                                                                                    \\
\midrule
Text                      & -     & -                                                                                  & 40K            & ShareGPT~\cite{sharegpt}                                                                                                                                                                                                                                                                          \\
\midrule
\textbf{Total}            & \textbf{1005M} &                                                                                    & \textbf{2006K} & -                                                                                                                                                                                               
\\
\bottomrule
\end{tabular}%
}
\caption{\textbf{Training Datasets.}}
\label{tab:appendix_data}
\end{table}

\subsection{Multimodal Instruction Tuning Dataset}

We summary the multimodal instruction tuning dataset in Tab.~\ref{tab:appendix_data}.

\subsection{Prompt Design}
\label{subsec:prompt_design}

The prompt formats for each dataset are shown in Tab.~\ref{tab:appendix_prompt_design}.

\begin{table}[h]
\centering
\resizebox{\linewidth}{!}{%
\begin{tabular}{ll}
\toprule
Dataset                                                                          & Prompt Format                                                                            \\
\midrule
\makecell[l]{LLaVA-150K~\cite{llava1}\\ShareGPT~\cite{sharegpt}\\Video Conversation~\cite{zhao2023chatbridge}\\Audio Conversation~\cite{zhao2023chatbridge}\\Point Conversation~\cite{xu2023pointllm}} & (use their original prompt)                                                              \\
\midrule
\makecell[l]{VQAv2~\cite{goyal2017vqav2}, GQA~\cite{hudson2019gqa}\\OKVQA~\cite{schwenk2022okvqa}\\OCRVQA~\cite{mishra2019ocrvqa}\\MSRVTT-QA~\cite{msvd}}                                             & \makecell[l]{\{Question\} Answer the question using a single\\ word or phase.}                           \\
\midrule
A-OKVQA~\cite{schwenk2022okvqa}                                                                          & \makecell[l]{\{Question\} \{Options\} Answer with the option's \\letter from the given choices directly} \\
\midrule
\makecell[l]{TextCaps~\cite{sidorov2020textcaps}\\COCO Caption~\cite{cococap}\\MSRVTT-Cap~\cite{xu2016msrvtt}\\AudioCaps~\cite{kim2019audiocaps}}                                    & \makecell[l]{Provide a one-sentence caption for the provided\\ image/video/audio.}                       \\
\midrule
\makecell[l]{RefCOCO~\cite{kazemzadeh2014referitgame}\\Visual Genome~\cite{krishna2017visual}}                                                           & Provide a short description for this region.                                             \\
\midrule
Ego4D~\cite{grauman2022ego4d}                                                                            & Describe the motion.                                                                     \\
\midrule
NSD~\cite{nsd}                                                                              & Describe the scene based on fMRI data.                                                  \\
\bottomrule
\end{tabular}%
}
\caption{\textbf{Prompt Formats for Training.}}
\label{tab:appendix_prompt_design}
\end{table}


\section{Evaluation Details}
\label{sec:eval_details}

\begin{table}[ht]
\centering
\resizebox{\linewidth}{!}{%
\begin{tabular}{ll}
\toprule
Dataset                                           & Prompt Format                                                                                                                                \\
\midrule
MMVet                                             & (use the original prompt)                                                                                                                    \\
\midrule
\makecell[l]{GQA~\cite{hudson2019gqa}\\VQAv2~\cite{goyal2017vqav2}\\OKVQA~\cite{okvqa}\\TextVQA~\cite{singh2019textvqa}\\MME~\cite{fu2023mme}\\MSVD~\cite{msvd}\\Clotho AQA~\cite{lipping2022clotho}\\MUSIC-AVQA~\cite{li2022musicvavqa}} & \makecell[l]{\{Question\} Answer the question using a single word or \\phase.}                                                               \\
\midrule
\makecell[l]{ScienceQA~\cite{lu2022learn}\\MMbench~\cite{liu2023mmbench}\\SEED-Bench~\cite{li2023seed}\\NextQA~\cite{nextqa}\\How2QA~\cite{li2020hero}}     & \makecell[l]{\{Question\} \{Options\} Answer with the option's letter from \\the given choices directly}                                     \\
\midrule
VizWiz~\cite{gurari2018vizwiz}                                            & \makecell[l]{\{Question\} When the provided information is insufficient, \\respond with 'Unanswerable'. Answer the question using \\a single word or phase.} \\
\midrule
\makecell[l]{Nocaps~\cite{agrawal2019nocaps}\\Flickr30K~\cite{plummer2015flickr30k}\\VATEX~\cite{wang2019vatex}\\VALOR~\cite{chen2023valor}\\Clotho Cap~\cite{drossos2020clotho}\\Objaverse-Cap~\cite{deitke2023objaverse}}    & \makecell[l]{Provide a one-sentence caption for the provided \\image/video/audio/point cloud.}                                                                                       \\
\midrule
AVSD~\cite{alamri2019audiovisual}                                              & \makecell[l]{\{Question\} Answer the question and explain the reason \\in one sentence.}                                                                      \\
\midrule
Objaverse-CLS~\cite{deitke2023objaverse}                                     & What is this?                                                                                                                                \\
\midrule
\makecell[l]{NYUv2~\cite{nyuv2}\\SUN RGB-D~\cite{song2015sun}}                                  & \makecell[l]{\{Class List\} What is the category of this scene? \\Choice one class from the class sets.}                                              \\
\bottomrule
\end{tabular}%
}
\caption{\textbf{Prompt Formats for Evaluation.}}
\label{tab:appendix_eval_prompts}
\end{table}

\begin{table*}[ht]
\centering
\resizebox{\textwidth}{!}{%
\small
\begin{tabular}{l|ccc|cccccccc}
\toprule
\multirow{2}{*}{Model} & \multirow{2}{*}{\makecell[c]{Encoder\\Param}} & \multirow{2}{*}{\#Encoder} & \multirow{2}{*}{\#Projection} & \multicolumn{8}{c}{Supported Modalities}                                                      \\
                       &                                &                            &                               & Image     & Video     & Audio     & Point     & IMU       & Depth     & Normal    & fMRI      \\
\midrule
X-LLM~\cite{chen2023xllm}                  & -                              & 3                          & 3                             & \ding{51} & \ding{51} & \ding{51} &           &           &           &           &           \\
PandaGPT~\cite{su2023pandagpt}               & 1.2B                           & 2                          & 1                             & \ding{51} & \ding{51} & \ding{51} &           &           &           &           &           \\
ImageBind-LLM~\cite{han2023imagebind}          & 1.8B                           & 3                          & 1                             & \ding{51} & \ding{51} & \ding{51} & \ding{51} &           &           &           &           \\
ChatBridge~\cite{zhao2023chatbridge}             & 1.3B                           & 3                          & 3                             & \ding{51} & \ding{51} & \ding{51} &           &           &           &           &           \\
AnyMAL~\cite{moon2023anymal}                 & 2B                             & 3                          & 3                             & \ding{51} & \ding{51} & \ding{51} &           & \ding{51} &           &           &           \\
\rowcolor{Gray}
\textbf{OneLLM (Ours)}          & \textbf{0.6B}                           & \textbf{1}                          & \textbf{1}                             & \ding{51} & \ding{51} & \ding{51} & \ding{51} & \ding{51} & \ding{51} & \ding{51} & \ding{51} \\
\bottomrule
\end{tabular}%
}
\caption{\textbf{Comparisons of Different Multimodal LLMs.}}
\label{tab:appendix_model_comparisons}
\end{table*}

In this section, we first list the evaluation prompts for each dataset in Tab.~\ref{tab:appendix_eval_prompts}. Then we will give more evaluation details.

\paragraph{Image, Video and Audio Tasks.} We evaluate all datasets using their official evaluation protocols. As shown in Tab.~\ref{tab:appendix_eval_prompts}, for QA tasks with options, we ask OneLLM to directly predict the option letters; For open-ended QA tasks, we ask OneLLM to predict a single word or phase. For captioning tasks, we ask OneLLM to generate a one-sentence caption. Note that for audio-video-text tasks, the input sequence to the LLM is: \textit{\{Video Tokens\} \{Audio Tokens\} \{Text Prompts\}}.

\paragraph{Point Cloud Tasks.} Our evaluation on point cloud tasks mainly follows PointLLM~\cite{xu2023pointllm}. For the point cloud classification task, we use the same prompt as PointLLM: \textit{What is this}, and evaluate the accuracy using GPT4.

\paragraph{Depth/Normal Map Tasks.} For scene classification using depth/normal map, we first prepend the category list to the beginning of prompt, then we ask OneLLM to choose one class for the list.

\paragraph{IMU/fMRI Tasks.} We evaluate on IMU/fMRI captioning tasks. The prompts are the same as their training prompts: \textit{Describe the motion} for IMU captioning and \textit{Describe the scene based on fMRI data} for fMRI captioning.

\section{Comparison with Prior Works}
\label{sec:comp}
The main difference between OneLLM and previous MLLMs is that we show a unified encoder is \textit{sufficient} to align multi-modalities with LLMs. As shown in Tab.~\ref{tab:appendix_model_comparisons}, OneLLM with \textit{one} universal encoder, \textit{on}e projection module and \textit{less} parameters (0.6B) can unify more modalities into one framework. The results in the main paper (Tab.1-6) also demonstrate that OneLLM can achieves better performance to previous works. The ablation experiments in Tab.7 (a) also show that jointly training all modalities with our unified framework can benefit data-scarce modalities. Here we are not trying to prove that OneLLM's architecture is optimal, but to show the possibility of building MLLMs using a unified and scalable framework.

\begin{figure*}[h]
    \centering
    \includegraphics[width=\textwidth]{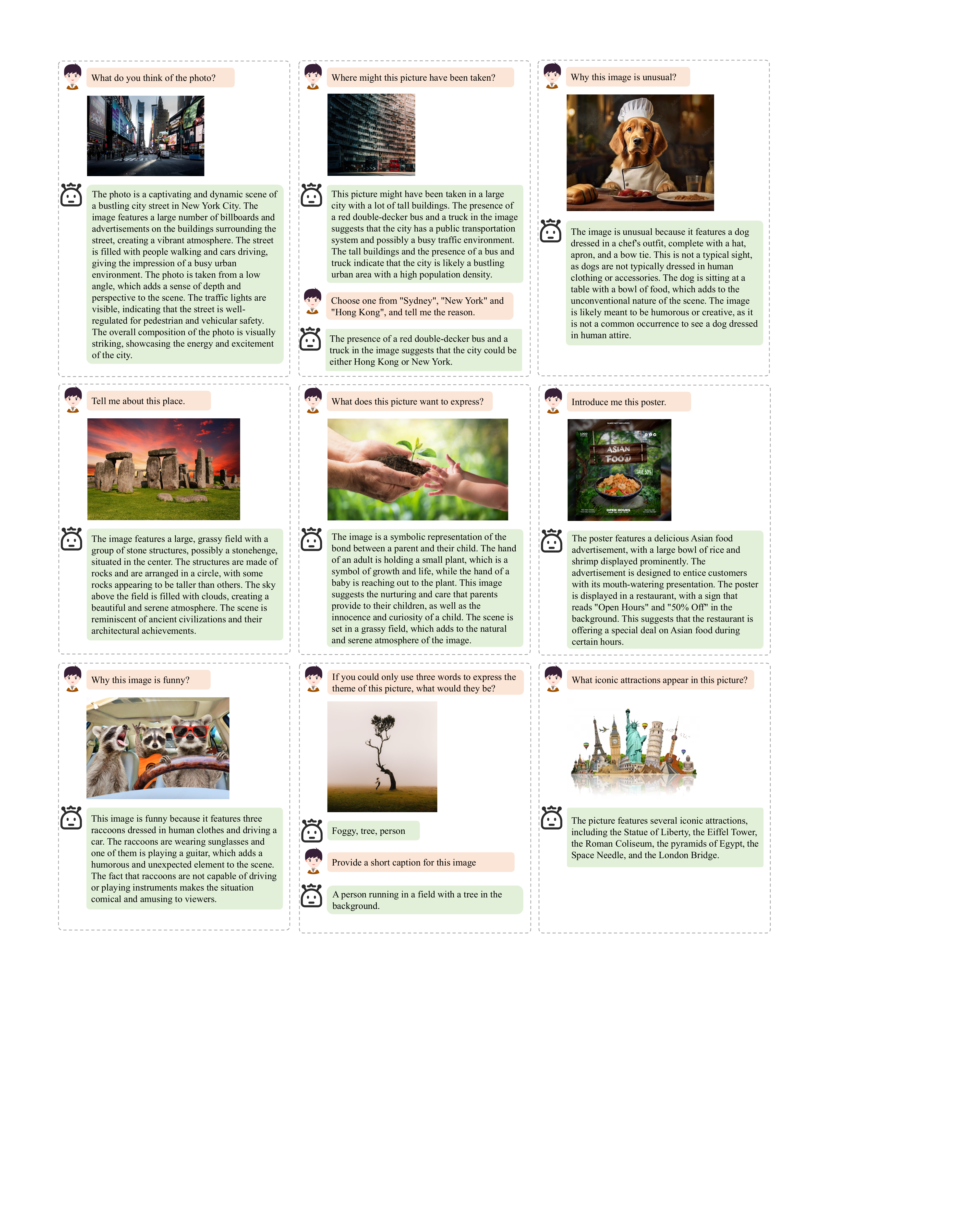}
    \caption{\textbf{Additional Qualitative Image Demos.}}
    \label{fig:appendix_demo_1}
\end{figure*}

\begin{figure*}[p]
    \centering
    \includegraphics[width=0.93\textwidth]{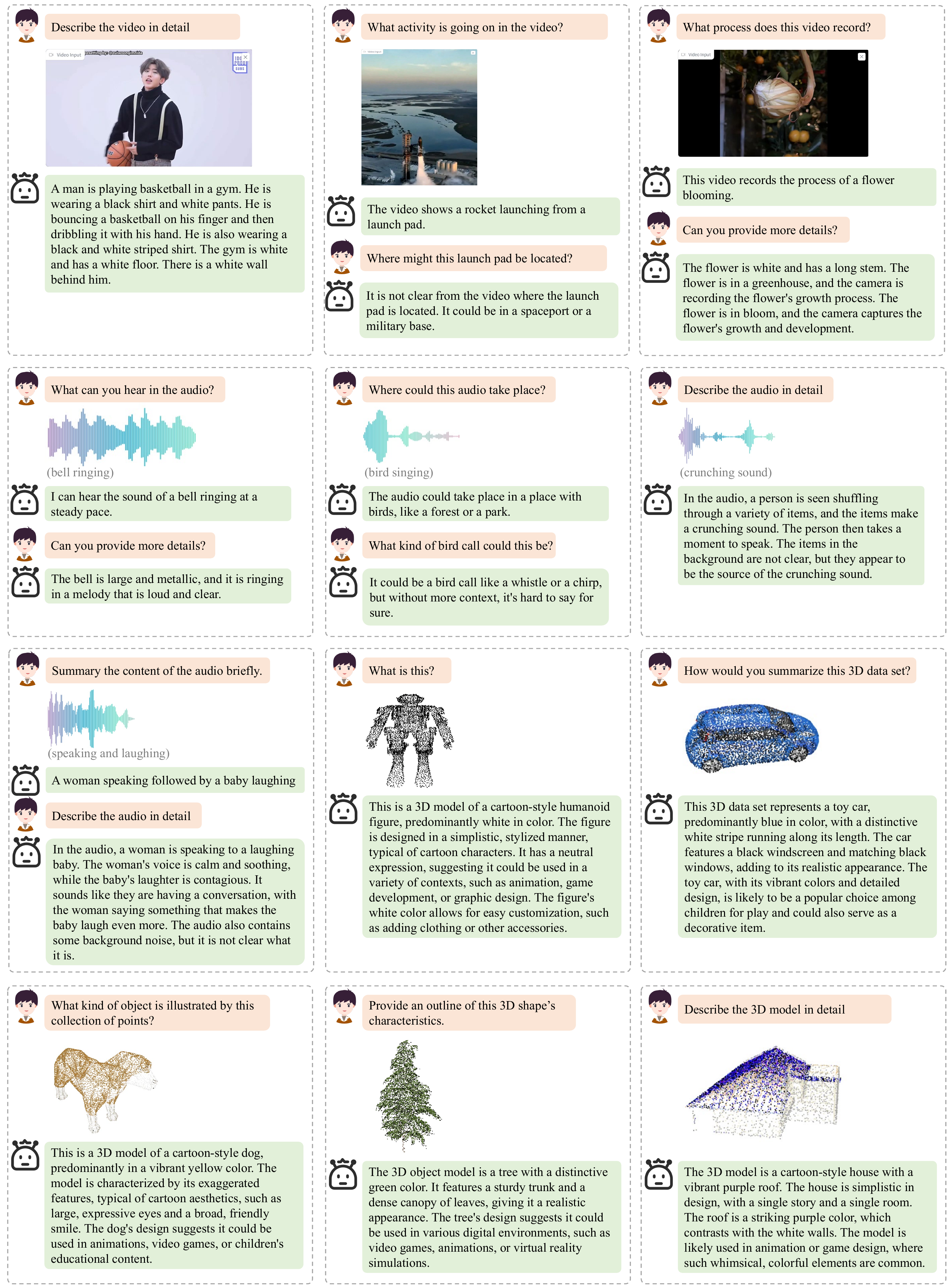}
    \caption{\textbf{Additional Qualitative Video, Audio and Point Cloud Demos.}}
    \label{fig:appendix_demo_2}
\end{figure*}

\begin{figure*}[!h]
    \centering
    \includegraphics[width=\textwidth]{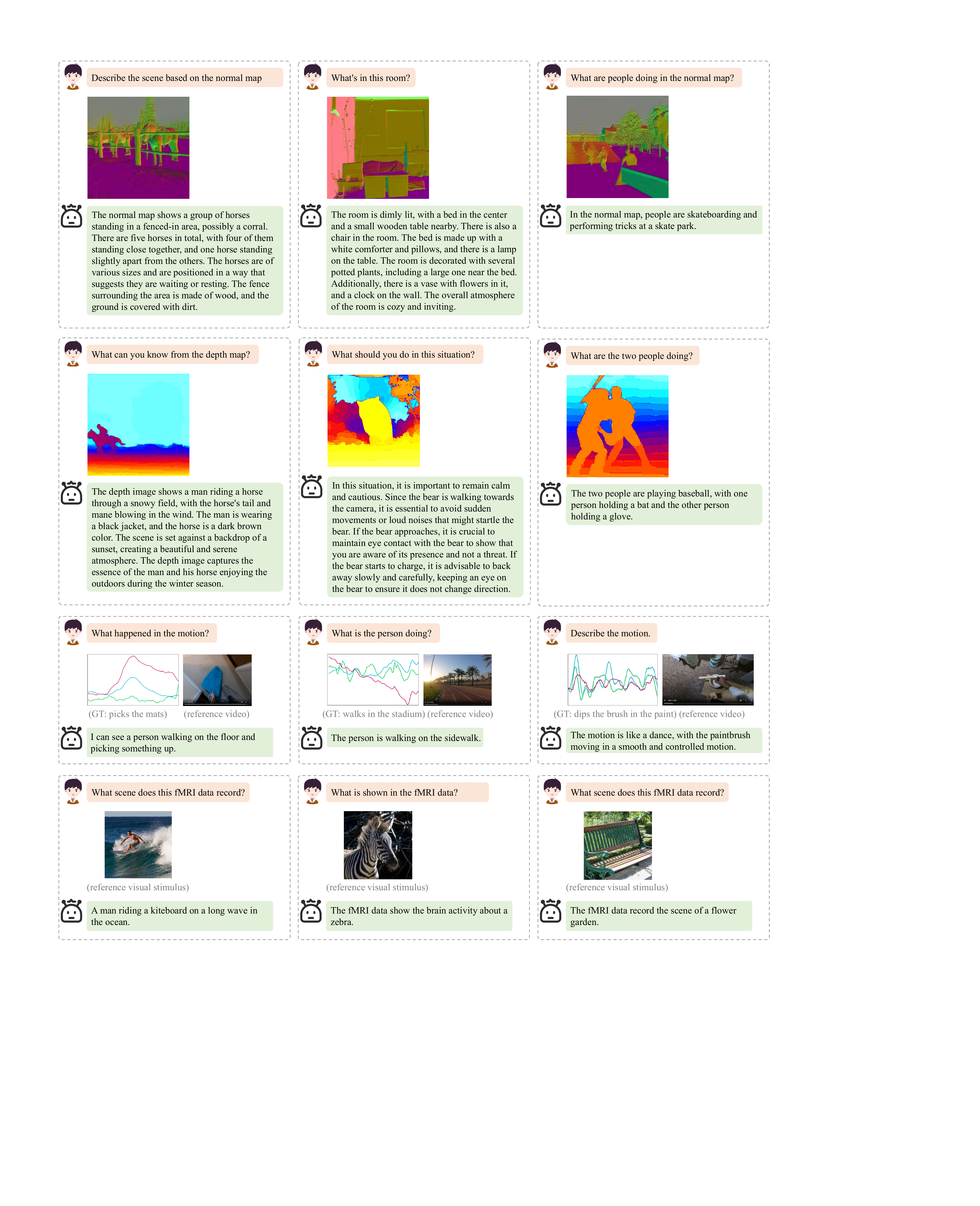}
    \caption{\textbf{Additional Qualitative Depth/Normal Map, IMU and fMRI Demos.}}
    \label{fig:appendix_demo_3}
\end{figure*}

\section{Additional Qualitative Results}
\label{sec:qualitative}

In this section, we provide more qualitative results in Fig.~\ref{fig:appendix_demo_1}, Fig.~\ref{fig:appendix_demo_2} and Fig.~\ref{fig:appendix_demo_3}.

{
    \small
    \bibliographystyle{ieeenat_fullname}
    \bibliography{main}
}


\end{document}

%% file: preamble.tex
\usepackage[dvipsnames]{xcolor}


%% file: tables/image_eval.tex
\begin{table*}[t]
\centering
\resizebox{\textwidth}{!}{%
\begin{tabular}{l|l|cccccc|cc|cccc}
\toprule
\multirow{2}{*}{Model}                        & \multirow{2}{*}{LLM} & \multicolumn{6}{c|}{VQA}                     & \multicolumn{2}{c|}{Image Caption} & \multicolumn{4}{c}{MM Benchmark} \\
                                              &                      & GQA   & VQAv2 & OKVQA & TVQA & SQA  & Vizwiz & NoCaps          & Flickr           & MME    & MMB  & MMVet & SEED \\
\midrule
\multicolumn{14}{c}{vision specialized LLM}                                                                                                                                 \\
\midrule
Flamingo-9B~\cite{alayrac2022flamingo}        & Chinchilla-7B        & -     & 51.8  & 44.7  & 30.1 & -    & 28.8   & -               & 61.5             & -      & -    & -     & -    \\
Flamingo-80B~\cite{alayrac2022flamingo}       & Chinchilla-70B       & -     & 56.3  & 50.6  & 31.8 & -    & 31.6   & -               & 67.2             & -      & -    & -     & -    \\
BLIP-2~\cite{li2023blip}                      & Vicuna-7B            & -     & -     & -     & 40.1 & 53.8 & -      & 107.5           & 74.9             & -      & -    & -     & -    \\
BLIP-2~\cite{li2023blip}                      & Vicuna-13B           & 41.0  & 41.0  & -     & 42.5 & 61   & 19.6   & 103.9           & 71.6             & 1293.8 & -    & 22.4  & -    \\
InstructBLIP~\cite{instructblip}              & Vicuna-7B            & 49.2  & -     & -     & 50.1 & 60.5 & 34.5   & 123.1           & 82.4             & -      & 36   & 26.2  & - \\
InstructBLIP~\cite{instructblip}              & Vicuna-13B           & 49.5  & -     & -     & 50.7 & 63.1 & 34.3   & 121.9           & 82.8             & 1212.8 & -    & 25.6  & - \\
IDEFICS-9B~\cite{laurenccon2023obelisc}       & LLaMA-7B             & 38.4  & 50.9  & 38.4  & 25.9 & -    & 35.5   & -               & 27.3             & -      & 48.2 & -     & -    \\
IDEFICS-80B~\cite{laurenccon2023obelisc}      & LLaMA-65B            & 45.2  & 60.0  & 45.2  & 30.9 & -    & 36.0   & -               & 53.7             & -      & 54.5 & -     & -    \\
LLaMA-Ad.v2~\cite{gao2023llama}               & LLaMA-7B             & 43.9  & -     & 55.9  & 43.8 & 54.2 & -      & 42.7            & 30.5             & 972.7  & 38.9 & 31.4  & 32.7 \\
Qwen-VL~\cite{bai2023qwen}                    & Qwen-7B              & 57.5  & 78.2  & 56.6  & 61.5 & 68.2 & 38.9   & 120.2           & 81.0             & 1487.5 & 60.6 & -     & 58.2 \\
LLaVA-v1.5~\cite{liu2023improvedllava}        & Vicuna-7B            & 62.0  & 78.5  & -     & 58.2 & 66.8 & 50.0   & -               & -                & 1510.7 & 64.3 & 30.5  & 58.6 \\
\midrule 
\multicolumn{14}{c}{multimodal generalist LLM}                                                                                                                             \\
\midrule
ImageBind-LLM~\cite{han2023imagebind}         & LLaMA-7B             & 41.1  & -     & -     & 24.0 & 51.4 & -      & 29.6            & 23.5             & 775.7  & -    & -     & -    \\
ChatBridge-13B~\cite{zhao2023chatbridge}      & Vicuna-13B           & \underline{41.8}  & -     & \underline{45.2}  & -    & -    & -      & \underline{115.7}           & \textbf{82.5}             & -      & -    & -     & -    \\
AnyMAL-13B~\cite{moon2023anymal}              & LLaMA2-13B           & -     & 59.6  & 33.1  & 24.7 & 52.7 & 24.4   & -               & -                & -      & -    & -     & -    \\
AnyMAL-70B~\cite{moon2023anymal}              & LLaMA2-70B           & -     & \underline{64.2}  & 42.6  & \underline{32.9} & \textbf{70.8} & \underline{33.8}   & -               & -                & -      & -    & -     & -    \\
\rowcolor{Gray}
\textbf{OneLLM-7B (Ours)}                     & \textbf{LLaMA2-7B}   & \textbf{59.5}  & \textbf{71.6}  & \textbf{58.9}  & \textbf{34.0} & \underline{63.4} & \textbf{45.9}   & \textbf{115.9}           & \underline{78.6}             & 1392.0 & 60.0 & 29.1  & \textbf{61.2} \\
\bottomrule
\end{tabular}%
}
\vspace{-3mm}
\caption{\textbf{Evaluation on 12 Image-Text Benchmarks}, including 6 VQA tasks (GQA~\cite{hudson2019gqa}, VQAv2~\cite{goyal2017vqav2}, OKVQA~\cite{okvqa}, TextVQA (TVQA)~\cite{singh2019textvqa}, ScienceQA (SQA)~\cite{lu2022learn} and Vizwiz~\cite{gurari2018vizwiz}), 2 image captioning tasks (Nocaps~\cite{agrawal2019nocaps} and Flickr30K~\cite{plummer2015flickr30k}), and 4 multimodal benchmarks (MME~\cite{fu2023mme}, MM Bench (MMB)~\cite{liu2023mmbench}, MMVet~\cite{yu2023mmvet} and SEED~\cite{li2023seed}). The LLMs are Chinchilla~\cite{hoffmann2022training}, Vicuna~\cite{vicuna}, Qwen~\cite{bai2023qwen_llm}, LLaMA~\cite{touvron2023llama} and LLaMA2~\cite{touvron2023llama2}. The evaluation metrics for VQA and captioning tasks are accuracy and CIDEr, respectively.
The results in \textbf{bold} and \underline{underline} are the best and second-best results, respectively. -: Not reported result.
}
\vspace{-3mm}
\label{tab:image_eval}
\end{table*}

%% file: tables/video_eval.tex
\begin{table}[t]
\centering
\resizebox{\linewidth}{!}{%
\begin{tabular}{lccccc}
\toprule
\multirow{2}{*}{Model} & \multirow{2}{*}{0-shot}      & NextQA          & How2QA           & MSVD      & VATEX                        \\
                                          &           & Acc.            & Acc.             & Acc.             & CIDEr     \\
\midrule
HGQA~\cite{xiao2022video}                 & \ding{55} & 51.8            & -                & 41.2              & -         \\
JustAsk~\cite{yang2021just}               & \ding{55} & 52.3            & 84.4             & 46.3              & -         \\
VALOR ~\cite{chen2023valor}               & \ding{55} & -               & -                & 60.0              & 95.1      \\
SeViLA~\cite{yu2023self}                  & \ding{55} & 73.8            & 83.6             & -                 & -         \\
\midrule
FrozenBiLM~\cite{yang2022zero}            & \ding{51} & -               & 58.4             & 33.8              & -         \\
InternVideo~\cite{wang2022internvideo}    & \ding{51} & \underline{49.1}& \underline{62.2} & \underline{55.5}  & -         \\
ChatBridge-13B~\cite{zhao2023chatbridge}  & \ding{51} & -               & -                & 45.3              & \textbf{48.9}      \\
AnyMAL-13B~\cite{moon2023anymal}          & \ding{51} & 47.9            & 59.6             & -                 & -         \\
\rowcolor{Gray}
\textbf{OneLLM-7B (Ours)}                 & \ding{51} & \textbf{57.3}   & \textbf{65.7}    & \textbf{56.5}     & \underline{43.8}      \\
\bottomrule
\end{tabular}%
}
\vspace{-3mm}
\caption{\textbf{Evaluation on Video-Text Tasks}, including video question answering (NextQA~\cite{nextqa}, How2QA~\cite{li2020hero} and MSVD~\cite{msvd}) and video captioning tasks (VATEX~\cite{wang2019vatex}). Acc.: Accuracy.}
\vspace{-2mm}
\label{tab:video_eval}
\end{table}

%% file: tables/audio_eval.tex
\begin{table}[t]
\centering
\resizebox{\linewidth}{!}{%
\small
\begin{tabular}{lcccc}
\toprule
\multirow{2}{*}{Model}              & \multirow{2}{*}{0-shot}    & \multicolumn{2}{c}{Clotho Caption} & Clotho AQA \\
                                    &                            & CIDEr           & SPIDEr           & Acc.       \\
\midrule
FeatureCut~\cite{ye2022featurecut}  & \ding{55}                  & 43.6            & 27.9             & -          \\
Wavcaps~\cite{mei2023wavcaps}       & \ding{55}                  & 48.8            & 31.0             & -          \\
MWAFM~\cite{li2023multi}            & \ding{55}                  & -               & -                & 22.2       \\
Pengi~\cite{deshmukh2023pengi}      & \ding{55}                  & -               & 27.1             & 64.5       \\
\midrule
LTU-7B~\cite{gong2023listen}        & \ding{51}                  & -               & 11.9             &            \\
ChatBridge-13B~\cite{zhao2023chatbridge}& \ding{51}              & 26.2            & -                & -          \\
\rowcolor{Gray}
\textbf{OneLLM-7B (Ours)}           & \ding{51}                  & \textbf{29.1}   & \textbf{19.5}             & \textbf{57.9}       \\
\bottomrule
\end{tabular}%
}
\vspace{-3mm}
\caption{\textbf{Evaluation on Audio-Text Tasks}, including audio captioning on Clotho Caption~\cite{drossos2020clotho} and audio question answering on Clotho AQA~\cite{lipping2022clotho}.}
\vspace{-5mm}
\label{tab:audio_eval}
\end{table}

%% file: tables/audio_visual_eval.tex
\begin{table}[t]
\centering
\resizebox{\linewidth}{!}{%
\small
\begin{tabular}{lcccc}
\toprule
\multirow{2}{*}{Model} & \multirow{2}{*}{0-shot}              & MUSIC-AVQA         &VALOR             & AVSD         \\
                       &                                      & Acc.               &CIDEr             & CIDEr         \\
\midrule
MAVQA~\cite{li2022musicvavqa}                  & \ding{55}    & 71.5               & -                & -             \\
VALOR~\cite{chen2023valor}                     & \ding{55}    & 78.9               & 61.5             & -             \\
VAST~\cite{chen2023vast}                       & \ding{55}    & 80.7               & 62.2             & -             \\
FA+HRED~\cite{nguyen2018film}                  & \ding{55}    & -                  & -                & 84.3          \\
MTN~\cite{le2019multimodal}                    & \ding{55}    & -                  & -                & 98.5          \\
COST~\cite{pham2022video}                      & \ding{55}    & -                  & -                & 108.5         \\
\midrule
ChatBridge-13B~\cite{zhao2023chatbridge}       & \ding{51}    & \underline{43.0}   & \underline{24.7} & \textbf{75.4}          \\
\rowcolor{Gray}
\textbf{OneLLM-7B (Ours)}                      & \ding{51}    & \textbf{47.6}      & \textbf{29.2}    & \underline{74.5}          \\
\bottomrule
\end{tabular}%
}
\vspace{-3mm}
\caption{\textbf{Evaluation on Audio-Video-Text Tasks}, including audio-visual question answering on MUSIC-AVQA~\cite{li2022musicvavqa} and audio-visual captioning on VALOR-32K~\cite{chen2023valor} and dialog completion on AVSD~\cite{alamri2019audiovisual}.}
\vspace{-2mm}
\label{tab:audio_visual_eval}
\end{table}

%% file: tables/point_eval.tex

\begin{table}[t]
\centering
\resizebox{\linewidth}{!}{%
\begin{tabular}{l|ccc|c}
\toprule
\multirow{2}{*}{Model}               & \multicolumn{3}{c|}{Captioning} & Classification \\
                                     & BLEU-1   & ROUGE-L   & METEOR  & GPT4-Acc.    \\
\midrule
InstructBLIP-7B~\cite{instructblip}  & 11.2     & 13.9      & 14.9    & 38.5           \\
InstructBLIP-13B~\cite{instructblip} & 12.6     & 15.0      & 16.0    & 35.5           \\
PointLLM-7B~\cite{xu2023pointllm}    & 8.0      & 11.1      & 15.2    & \textbf{47.5}           \\
PointLLM-13B~\cite{xu2023pointllm}   & 9.7      & 12.8      & 15.3    & 45.0           \\
\rowcolor{Gray}
\textbf{OneLLM-7B (Ours)}           & \textbf{42.2}     & \textbf{45.3}      & \textbf{20.3}    & 44.5           \\
\bottomrule
\end{tabular}%
}
\vspace{-3mm}
\caption{\textbf{Evaluation on Point Cloud-Text Tasks}. The evaluation dataset is from Objaverse~\cite{deitke2023objaverse}, following the data split in PointLLM~\cite{xu2023pointllm}. InstructBLIP takes single-view image as input, while PointLLM and OneLLM take point cloud as input. GPT4-Acc.: GPT4 as the accuracy evaluator~\cite{xu2023pointllm}.}
\vspace{-3mm}
\label{tab:point_eval}
\end{table}

%% file: tables/depth_normal_eval.tex
\begin{table}[t]
\centering
\small
\setlength\tabcolsep{8.2pt}
\begin{tabular}{lccc}
\toprule
\multirow{2}{*}{Model}                & \multirow{2}{*}{0-shot} & NYUv2 & SUN RGB-D \\
                                      &           & Acc.  & Acc.      \\
\midrule
ImageBind~\cite{girdhar2023imagebind} & \ding{55} & 54.0  & 35.1      \\
Omnivore~\cite{girdhar2022omnivore}   & \ding{55} & 76.7  & 64.9      \\
\midrule
Random                                & \ding{51} & 10.0  & 5.26      \\
CLIP ViT-H$^*$~\cite{radford2021learning} & \ding{51} & 41.9  & \underline{25.4}      \\
\rowcolor{Gray}
\textbf{OneLLM-N (Ours)}                       & \ding{51} & \underline{46.5}  & 21.2      \\
\rowcolor{Gray}
\textbf{OneLLM-D (Ours)}                       & \ding{51} & \textbf{50.9}  & \textbf{29.0}      \\
\bottomrule
\end{tabular}%
\vspace{-3mm}
\caption{\textbf{Evaluation on Scene Classification Tasks Using Depth / Normal Map.} OneLLM-N/D: OneLLM with Depth / Normal map inputs. Note that NYUv2~\cite{nyuv2} and SUN RGB-D~\cite{song2015sun} only have depth maps, we adopt pretrained DPT model~\cite{eftekhar2021omnidata} to generate normal maps. $^*$: The input to CLIP is depth rendered grayscale image. ImageBind is trained on image-depth pairs of SUN RGB-D and therefore is not zero-shot.}
\vspace{-5mm}
\label{tab:depth_normal_eval}
\end{table}

%% file: tables/ablation.tex
\begin{table}[t]
\centering
\resizebox{\linewidth}{!}{%
\begin{tabular}{lllll}
\toprule
Task              & NoCaps & VQAv2 & ClothoQA & MSVDQA \\
\midrule
\multicolumn{5}{l}{\bf (a) Training Mode}             \\
\midrule
Separate          & 115.6{\footnotesize(\textcolor{Red}{-0.2})}  & 71.9{\footnotesize(\textcolor{Green}{+0.3})}      & 37.8{\footnotesize(\textcolor{Red}{-19.6})}     &  31.0{\footnotesize(\textcolor{Red}{-25.8})}      \\
\rowcolor{Gray}
Joint       & 115.8  & 71.6  & 57.4     & 56.8   \\
\midrule
\multicolumn{5}{l}{\bf (b) Weight Initialization}      \\
\midrule
Random Init.     & 98.8{\footnotesize(\textcolor{Red}{-17.0})}& 65.6{\footnotesize(\textcolor{Red}{-6.0})}  & 57.6{\footnotesize(\textcolor{Green}{+0.2})}     & 53.1{\footnotesize(\textcolor{Red}{-3.7})}   \\
\rowcolor{Gray}
Image Init.      & 115.8  & 71.6  & 57.4     & 56.8   \\
\midrule
\multicolumn{5}{l}{\bf (c) Number of Experts (Parameters)}         \\
\midrule
1 {\small(88M)}            & 108.7{\footnotesize(\textcolor{Red}{-7.1})}  & 66.9{\footnotesize(\textcolor{Red}{-4.7})}  & 58.2{\footnotesize(\textcolor{Green}{+0.8})}     & 53.3{\footnotesize(\textcolor{Red}{-3.5})}   \\
\rowcolor{Gray}
3 {\small(264M)}           & 115.8  & 71.6  & 57.4     & 56.8   \\
5 {\small(440M)}           & 114.6  & 71.7  & 58.2     & 56.7   \\
7 {\small(616M)}           & 114.9  & 71.6  & 58.8     & 56.0   \\
\midrule
\multicolumn{5}{l}{\bf (d) Router Type}               \\
\midrule
Constant Router   & 109.8{\footnotesize(\textcolor{Red}{-6.0})}  & 67.7{\footnotesize(\textcolor{Red}{-3.9})}  & 56.2{\footnotesize(\textcolor{Red}{-1.2})}         & 55.3{\footnotesize(\textcolor{Red}{-1.5})}       \\
Sparse Router     & 112.8{\footnotesize(\textcolor{Red}{-3.0})}  & 71.1{\footnotesize(\textcolor{Red}{-0.5})}      & 56.7{\footnotesize(\textcolor{Red}{-0.7})}     & 55.7{\footnotesize(\textcolor{Red}{-1.1})}   \\
\rowcolor{Gray}
Soft Router       & 115.8  & 71.6  & 57.4     & 56.8   \\
\bottomrule
\end{tabular}%
}
\vspace{-3mm}
\caption{\textbf{Ablation Experiments.} We choose three modalities (image, audio, video) and four datasets (NoCaps~\cite{agrawal2019nocaps}, VQAv2~\cite{goyal2017vqav2}, ClothoQA~\cite{lipping2022clotho} and MSVDQA~\cite{msvd}) for evaluation. The row with \colorbox{Gray}{gray background} is our default setting.}
\vspace{-3mm}
\label{tab:ablation}
\end{table}